\documentclass[journal]{IEEEtran}%,draftcls,onecolumn
\ifCLASSINFOpdf
\else
 \usepackage[dvips]{graphicx}
\fi
\usepackage[varg]{txfonts}
\usepackage[tight,footnotesize]{subfigure}
\begin{document}
\title{Tracking Deformable Parts via Dynamic Conditional Random Fields}

\author{Suofei~Zhang, Zhixin~Sun, Xu~Cheng, and~Zhenyang~Wu,~\IEEEmembership{Member,~IEEE}
\thanks{S. Zhang and Z. Sun are with the School of Internet of Things, Nanjing University of Posts and Telecommunications, Nanjing, Jiangsu, 210003, China (e-mail: zhangsuofei@njupt.edu.cn; sunzx@njupt.edu.cn).}
\thanks{X. Cheng and Z. Wu are with the School of Information Science and Engineering, Southeast University, Nanjing, Jiangsu, 210096, China (e-mail: xcheng@seu.edu.cn; zhenyang@seu.edu.cn).}
\thanks{This work was supported by the Chinese National Natural Science Foundation (Grant No. 61373135, 60973140 and 61170276).}% <-this % stops a space
%\thanks{Manuscript received April 19, 2005; revised January 11, 2007.}
}


\markboth{Journal of \LaTeX\ Class Files,~Vol.~6, No.~1, January~2007}%
{Shell \MakeLowercase{\textit{et al.}}: Bare Demo of IEEEtran.cls for Journals}



\maketitle

\begin{abstract}
Despite the success of many advanced tracking methods in this area, tracking targets with drastic variation of appearance such as deformation, view change and partial occlusion in video sequences is still a challenge in practical applications. In this letter, we take these serious tracking problems into account simultaneously, proposing a dynamic graph based model to track object and its deformable parts at multiple resolutions. The method introduces well learned structural object detection models into object tracking applications as prior knowledge to deal with deformation and view change. Meanwhile, it explicitly formulates partial occlusion by integrating spatial potentials and temporal potentials with an unparameterized occlusion handling mechanism in the dynamic conditional random field framework. Empirical results demonstrate that the method outperforms state-of-the-art trackers on different challenging video sequences.
\end{abstract}


% \fi
\IEEEpeerreviewmaketitle

\section{Introduction}
\label{intro}


%\hfill January 11, 2007

\IEEEPARstart{V}{isual} tracking plays an essential role for many higher level understanding of video contents such like traffic surveillance, analysis of human behaviours and interactions between targets of interest, etc. During the past decade, some quite efficient object tracking methods~\cite{meanshift,isard} have been widely distributed in various applications. However, designing a robust tracking algorithm for realistic task is still a major challenge. The problems arise not only from intra-class variation of appearance caused by deformation and viewpoint change, but also from partial occlusion and cluttered background, etc.

For deformation and viewpoint change, recently, researchers tend to address the problem with online learning method to update the target model~\cite{5674053,ross2008incremental}. Such methods provide an effective way to handle universal tracking problems by achieving a synergy between tracking and recognition. However, for vast majority of common objects in daily life, e.g., pedestrians and vehicles, the object tracking by human eyes actually follows the recognition of target at the first glimpse. The leverage of massive experience in this recognition process brings high-level auxiliary knowledge to handle various problems in tracking. Motivated by this intuition, we propose to track objects via high performance object detection models, Deformable Part based Models (DPMs)~\cite{5255236}, in this letter. The similar inspiration also exists in other state-of-the-art work in the community~\cite{6302190}, although here we track the whole target as well as deformable parts simultaneously.
\begin{figure}[htbp]%[H]
  \centering
  \subfigure[]{
    \label{fig:cover-a}
    \includegraphics[height=4.2cm]{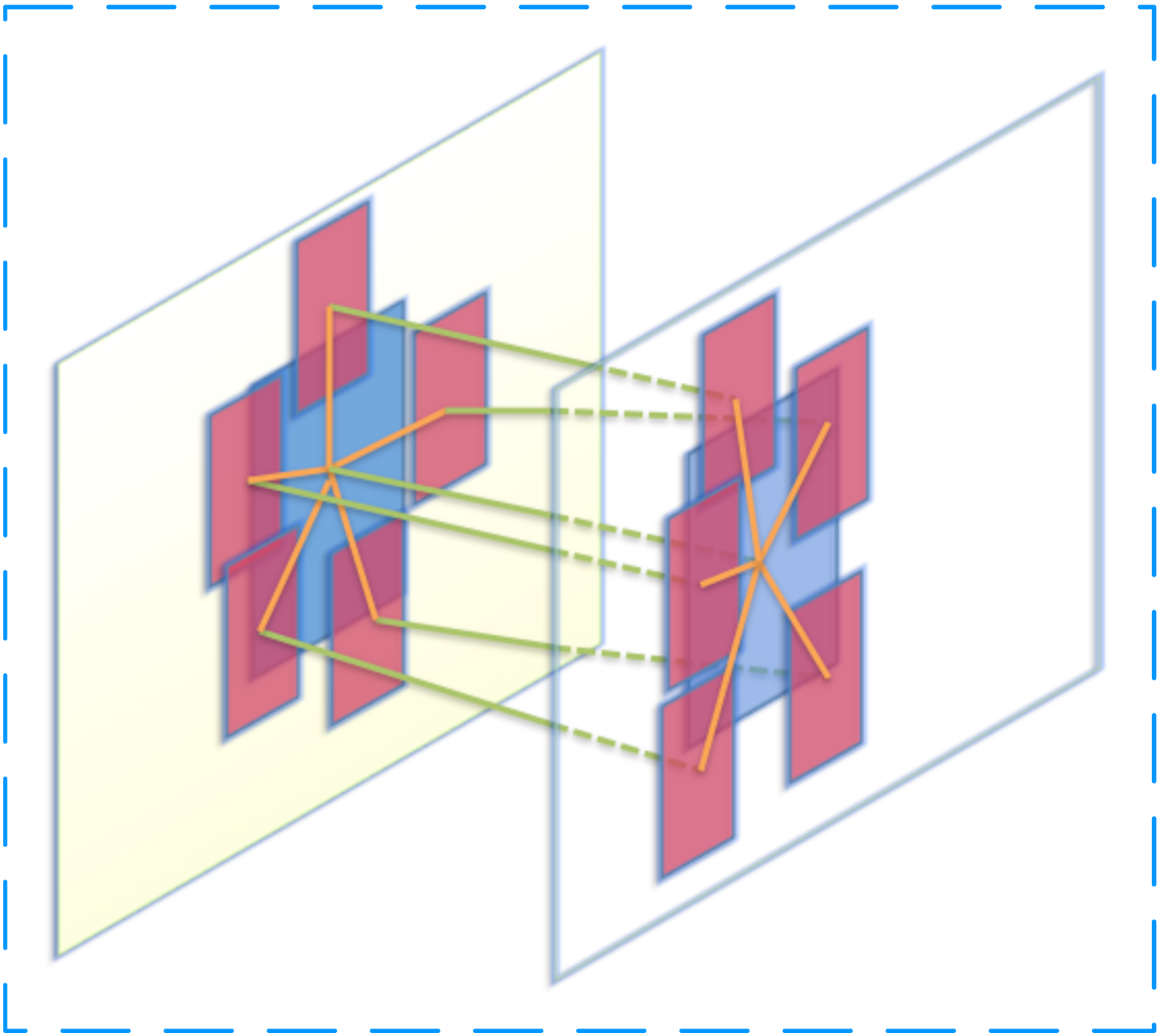}}
  \centering
  \subfigure[]{
    \label{fig:cover-b}
    \includegraphics[height=4.2cm]{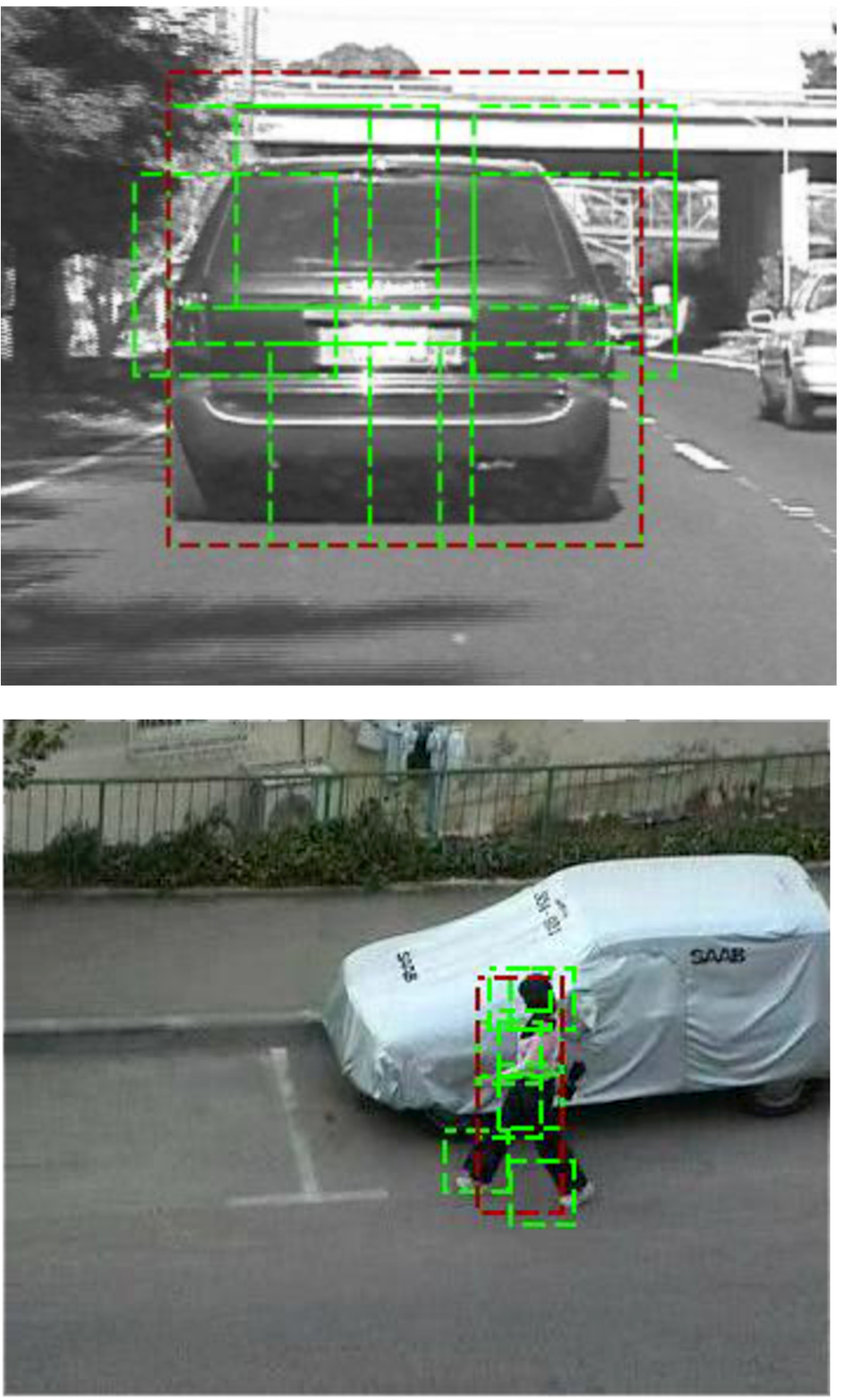}}
  \caption{(a) The graph based models with pre-defined spatial and temporal potentials between vertices over frames.  (b) Tracking results of pedestrian and car. Our method tracks not only the target, but also its detailed parts.}
  \label{fig:cover}
\end{figure}

The proposed tracking framework in this letter consists of several components which correspond to specific views of object. As shown in Figure.~\ref{fig:cover}, each component is a Dynamic Conditional Random Field (DCRF)~\cite{wang2} over consecutive frames to describe the details of objects on different resolutions. Each vertex in the graph is connected with its spatial and temporal neighbors by pre-defined pairwise potentials which  formulate the deformation of object. On bottom of that, a pyramid based representation of image effectively handle the illumination and scale change of target over frames.

For partial occlusion in cluttered background, part based models have yielded attractive results in recent progress of object tracking~\cite{1640835,6212358,6287549}.  A series of solutions attempt a sparse representation of objects~\cite{6212358,5740923} to track parts of target and thus handle partial occlusion problems. Differing from these decomposition based methods, our method can originally describe the status that some parts are absent from sight while a hypothesis of object is still credible due to other observed parts, and thus can handle occlusion more directly and flexibly.

The main contributions of this work are threefold: (1) we integrate high performance object detection method with dynamic graph based model, implementing an efficient object tracking with structured outputs; (2) we propose some novel temporal pairwise potentials to model the transition between parts over frames; and (3) we implement an efficient unparameterized logistic regression based mechanism, combining it with prior knowledge from previous frame to handle partial occlusion. Experiments on challenging video sequences prove the efficacy of our proposed method.

\section{Deformable Part Based Model}
DPM has been proven as quite effective model to formulate the significant intra-class variation of objects in challenging object detection problems. A representation of object by DPM can be considered as a mixture of $S$ star-shaped Conditional Random Fields (CRFs)~\cite{lafferty2001conditional} as components. Each component $s$ consists of one root part $x_0$ and $n$ deformable parts $(x_1,\ldots,x_n)$ as vertices of graph. The unary potential of vertex in $s$, which models the part appearance, is the output of Histogram of Oriented Gradients (HOG) features $\phi(H,x_j)$ filtered by a template function $F_{x_j}$, where $H$ is the HOG feature pyramid, $j\in{0,\ldots,n}$. The pairwise potential between root and part, which models the deformation, penalizes the displacement $v_{x_j}$ of part from the anchor position of trained model with a Gaussian kernel parameterized by a four-tuple $d_{x_j}$.

By considering $\phi(H,x_j)$s and $v_{x_j}$s as input, as well as $F_{x_j}$s and $d_{x_j}$s as parameters, we can realize the CRF output from the perspective of linear perceptron:
\begin{eqnarray}
  \label{eq:dpmc}
  \psi(H,c)&\hspace{-3mm}=\hspace{-3mm}&(F_{x_0},\ldots,F_{x_n},d_{x_0},\ldots,d_{x_n},b_c)^T\nonumber\\
  &\hspace{-3mm}&(\phi(H,x_0),\ldots,\phi(H,x_n),v_{x_0}\ldots,v_{x_n},1),
\end{eqnarray}
where $b_c$ is often termed as bias constant in this context. The correspondence between CRF and linear perceptron leads to a Support Vector Machine (SVM) based training framework in~\cite{5255236}. The efficacy of DPM arises from 3 building blocks: (1) the HOG pyramid handles the illumination and scale changes; (2) the mixture model takes multiple views of objects into account simultaneously; (3) The deformation penalty which is formulated by pairwise potentials in CRF tackles non-rigid deformations and intra-class variation in shape directly.

\section{Occlusion Handling}
Despite the great success that DPM has been witnessed, it has been reported that detecting partial occluded objects with DPM remains a challenging problem~\cite{6247879}. In this letter, we propose a similar but more efficient strategy to that of~\cite{6247879} to handle the partial occlusion problem. From Eq.~\ref{eq:dpmc}, one can see that in the CRF $s$ of DPM, the $score(x_j)$ related to each vertex can be computed separately as
\begin{equation}
  \label{eq:svert}
  score(x_j)=(F_{x_j},d_{x_j})^T(\phi(H,x_j),v_{x_j}).
\end{equation}

By using logistic regression over the SVM output $score(x_j)$ on every vertex~\cite{5995720}, we can model the probability of the hypothesis that a vertex appears at current site $s(x_j)$ as
\begin{equation}
  p(s(x_j)|F_{x_j},d_{x_j})=\frac{exp(score(x_j))}{1+exp(score(x_j))}.%=\frac{exp((F_j',d_j)^T(\phi(H,p_j),v_j))}{1+exp((F_j',d_j)^T(\phi(H,p_j),v_j))}
\end{equation}
If an object is partially occluded, aggregating the scores of all parts $X={x_0,\ldots,x_n}$ as in conventional DPM is apparently unsuitable. Therefore we only select a subset $X_c=\{x_k,\ldots,x_l\}$ of parts from $X$, finding the optimal $X_c^*$ to maximize the mean of normalized scores of vertices in $X_c$ as follows:
\begin{equation}
  \label{eq:maxsc}
  \psi'(H,c,X_c^*)=\max_{X_c}\frac{1}{|X_c|}\sum_{j\in{X_c}}p(s(x_j)|F_{x_j},d_{x_j}).
\end{equation}

\begin{figure}
  \centering
  \includegraphics[width=8.5cm]{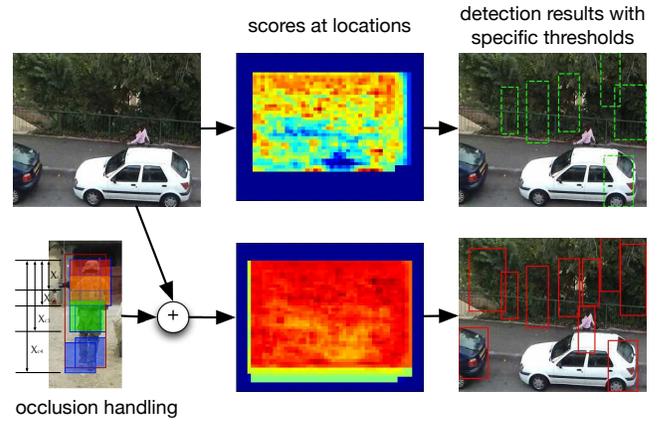}
  \caption{Difference between detecting with and without occlusion handling. By using unparameterized logistic regression and four recommended candidates of $X_c^*$, it is easy to observe that more noises along with possible correct hypotheses can be exploited by a rational threshold.}
  \label{fig:occhan}
\end{figure}

For common pedestrian tracking, similar to~\cite{6247879}, we only take four possible subsets of parts into consideration as in Figure~\ref{fig:occhan}. It has been proven that such limited choices are representative enough in most practical scenarios~\cite{6247879}. For more universal object tracking problems, a simple greedy search algorithm can be employed here to add parts into $X_c$ sequentially with trivial overhead of computation. Differing from the parameterized logistic regression in~\cite{6247879}, our method directly projects the output of SVM from $(-\infty,+\infty)$ to $(0,1)$ without any requirement of training stage. Such  simpler formulation is more flexible in various realistic tracking applications. From an empirical analysis as shown in Figure~\ref{fig:occhan}, our proposed occlusion handling strategy actually introduces noises into final detection results of DPM. However, it is still very promising since it compresses the distribution of DPM scores and allows some parts of object contribute to the result equally as the whole star-shaped model.

\section{Tracking via Dynamic Conditional Random Fields}
\subsection{Dynamic Conditional Random Field}
DCRF was proposed in~\cite{wang2} to implement an accurate foreground segmentation in video sequences. Here we introduce it into object tracking by integrating it with pre-defined potential functions from DPM. DCRF models the states of two random fields $s_t$ and $s_{t+1}$ over consecutive frames via Bayes' rule:
\begin{equation}
  \label{eq:bayes}
  p(s_{t+1}|o_{1:t+1})=\frac{1}{Z}p(o_{t+1}|s_{t+1})\sum_{s_t}p(s_{t+1}|s_t)p(s_t|o_{1:t}),
\end{equation}
where $Z$ is the partition function. Since $s_t$ indicates a random field which contains $|X|$ vertices here, to enumerate all possible states of $s_t$ in Eq.~\ref{eq:bayes} is intractable. Inspired by the derivation in~\cite{wang2}, we attempt to restrict the problem to every single vertex in $s$.

According to the Markov property and Hammersley-Clifford theorem, the state transition probability $p(s_{t+1}|s_t)$ in Eq.~\ref{eq:bayes} can be given by a Gibbs distribution as follows:
\begin{eqnarray}
  \label{eq:tempp}
  p(s_{t+1}|s_t)&\hspace{-3mm}\propto\hspace{-3mm}&\exp\bigg\{\sum_{x\in{X}}\big[V_x(s_{t+1}(x)|s_t(M_x))\nonumber\\
    &\hspace{-3mm}&+\sum_{y\in{N_x}}V_{x,y}(s_{t+1}(x),s_{t+1}(y))\big]\bigg\},
\end{eqnarray}
where $x$ and $y$ are vertices in the graph. The temporal neigborhood $M_x$ denotes the vertices at the $t$th frame which can impact $x$ at the $(t+1)$th frame, and the spatial neighbourhood $N_x$ refers to the spatially related vertices at the same frame to $x$. Here $s_t(M_x)$ stands for the state of neighboring vertex $\{s_t(y)|y\in{M_x}\}$, $V_x(\cdot)$ and $V_{x,y}(\cdot)$ are clique potentials related to the vertex $x$.

Due to the star shape of DPM, the adopted graph model in our proposed DCRF framework retains a facile structure. The posterior distribution $p(s_t|o_{1:t})$ for a site at the $t$th frame can be directly factorized as
\begin{equation}
  \label{eq:postf}
  p(s_t|o_{1:t})=\prod_{x\in{X}}p(s_t(x)|o_{1:t}(x)).
\end{equation}

With similar conditional independence assumption in~\cite{wang2}, the observation model $p(o_{t+1}|s_{t+1})$ the  can also be evaluated by product of likelihoods of vertices:
\begin{equation}
  \label{eq:likef}
  p(o_{t+1}|s_{t+1})=\prod_{x\in{X}}p(o_{t+1}(x)|s_{t+1}(x)).
\end{equation}

Combining Eq.~\ref{eq:bayes},~\ref{eq:tempp},~\ref{eq:postf} and~\ref{eq:likef} with Jensen's inequality, we can approximate the lower bound of $p(s_{t+1}|o_{1:t+1})$ at current site as
\begin{eqnarray}
  \label{eq:comp}
  &&\prod_{x\in{X}}\exp\bigg\{\big[\underbrace{V_x(o_{t+1}(x)|s_{t+1}(x))+\sum_{y\in{N_x}}V_{x,y}(s_{t+1}(x),s_{t+1}(y))}_{score(x)}\big]\nonumber\\
  &&\cdot\frac{1}{|M_x|}\sum_{y\in{M_x}}\sum_{s_t(y)}V_x(s_{t+1}(x)|s_t(y)p(s_t(y)|o_{1:t}(y)))\bigg\}.
\end{eqnarray}
Here we only consider corresponding vertex at previous frame as in Figure~\ref{fig:cover}, so $|M_x|=1$, $S_t(y)$ can be simply replaced by $S_t(x)$. As shown in Eq.~\ref{eq:comp}, the summation of unary potentials $V_x{\cdot}$ and pairwise potentials $V_{x,y}(\cdot)$ at every vertex corresponds to the output of DPM $score(x)$. Therefore, the equation has a very clear explanation: the local energy on a vertex of DCRF consists of DPM score as observation, temporal potential as transition function, and result from previous frame as posterior distribution. Since each vertex only has two possible states $s_t(x)\in\{0,1\}$, which indicate whether it occurs, Eq.~\ref{eq:comp} can be computed very efficiently especially in the logarithmic form.

\subsection{Temporal Potential Function}
To model the status that the object is partially occluded, we propose a novel transition function $V_x(s_{t+1}(x)|s_t(x))$ to impose the temporal connectivity between same parts over different frames. 
\begin{eqnarray}
\label{eq:tpf}
  V_x(s_{t+1}(x)|s_t(y))&\hspace{-3mm}=\hspace{-3mm}&\mathcal{G}(x-y;\Sigma)\cdot\delta(s_{t+1}(x)-s_t(y))\\%frac{e^{(x-y)'\Sigma^{-1}(x-y)}}{\sqrt{2\pi|\Sigma|}}
  &&+\frac{1}{1+e^{-||v_x-v_y||^2}}(1-\delta(s_{t+1}(x)-s_t(y))).\nonumber
\end{eqnarray}
The proposed temporal potential ensures the consistency between neighboring vertices. If the part $x$ is assumed to be observed at last frame, a normalised Gaussian kernel, $\mathcal{G}(x-y;\Sigma)$ is adopted to measure the motions of object. Here $\Sigma$ is a three-dimensional covariance matrix constraining the object into a relevant range on HOG feature pyramid. Otherwise if the part is assumed to be occluded, which means the direct prior knowledge about current part from last frame is absent, we keep the temporal connectivity with the difference of part deformation instead. It implies that if the pose of object changes drastically over frames, the final tracking result should be biased more on observation model rather than prior knowledge.

\section{Empirical Results}
We empirically testified the proposed graph model based tracking framework with three experiments. In experiments we adopted the DPMs trained for PASCAL VOC 2009~\cite{pascal-voc-2009}, which contain six components consisting of nine deformable parts. The algorithm is initialized by detecting the optimal window which overlaps with ground-truth by at least 70\% at the first frame. Only related HOG features at neighboring levels in pyramid are extracted for tracking. This configuration implies that the efficiency of our method is decided by both shown object size as well as image size. Zooming in frames of video directly will not bring any impact to the speed of tracking.

\subsection{From Detection to Tracking}
\begin{figure}
  \centering
  \includegraphics[width=8.5cm]{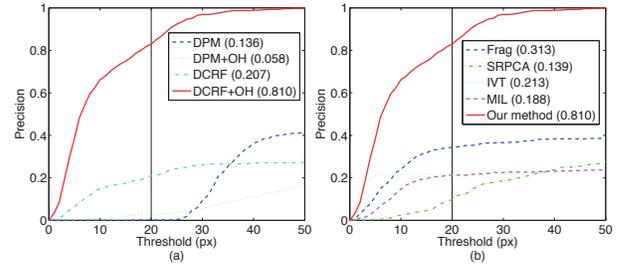}
  \caption{Quantitative comparisons of different tracking methods on ``Woman'' sequence: (a) performances of our method with different configurations, (b) performances of other leading methods and our proposed method.}
  \label{fig:exp1}
\end{figure}

Since tackling long-term partial occlusion is a main concern in this letter, we carefully evaluate the influences of proposed novel occlusion handling mechanisms in this section. A challenging video sequence, the ``Woman" sequence~\cite{1640835}, is used to evaluate the performances of four different configurations: detection by DPM directly (DPM), detection by DPM and occlusion handling (DPM+OH), tracking by DCRF merely with Gaussian kernel in Eq.~\ref{eq:tpf} (DCRF), and tracking with complete temporal potential function (DCRF+OH). Since there is no tracking failure problem for detection methods, we follow the evaluation protocol proposed by~\cite{5674053} in Figure~\ref{fig:exp1}.
%\begin{figure}%[H]
%  \centering
%  \subfigure[]{
%    \label{fig:exp1-a}
%    \includegraphics[width=4.3cm]{exp1.eps}}\hspace{-3mm}
%  \centering
%  \subfigure[]{
%    \label{fig:exp1-b}
%    \includegraphics[width=4.3cm]{exp1_2.eps}}
%  \caption{Comparisons of different tracking methods on ``Woman'' sequence: (a) performances of our method with different configurations, (b) performances of other leading methods and our proposed method.}
%  \label{fig:exp1}
%\end{figure}
 
It is easy to observe in Figure~\ref{fig:exp1}(a) that the proposed tracking method brought significant improvement to DPM based detection, despite that using unparameterized occlusion handling actually leads to worse result. Note that tracking with DCRF without occlusion handling achieved a desirable result at the beginning of the sequence. However the method failed to follow the target around frame \#125, where a long-term partial occlusion occurs, and finally leaded to mitigated result. An implementation with MATLAB on a Pentium 3.3 GHz CPU can process one frame in 0.7 second on this sequence, which is relatively much faster than detecting directly (2.5 second per frame).

\subsection{Comparison on ``Woman'' Sequence}
\begin{figure}%[htbp]
  \centering
  \includegraphics[width=9cm]{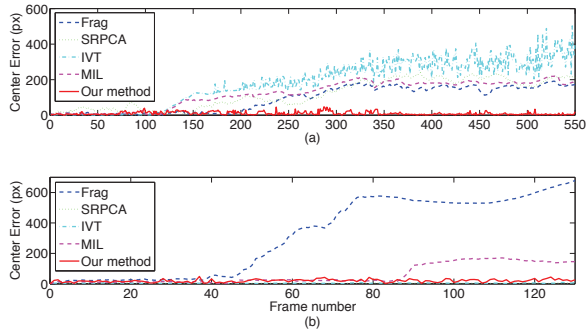}
  \caption{The performances of various methods on two video sequences measured by center errors. (a) ``Woman'' sequence, (b) ``Car 4'' sequence.}
  \label{fig:exp2}
\end{figure}

We also compared the performance of our proposed method with other leading tracking methods on the ``Woman'' Sequence. We took several representative state-of-the-art methods into account, i.e., Frag tracker~\cite{1640835}, SRPCA tracker~\cite{6212358}, IVT tracker~\cite{ross2008incremental} and MIL tracker~\cite{5674053}. It can be observed from Figure~\ref{fig:exp2}(a) that only our method successfully followed the pedestrian through the whole ``Woman'' sequence, while other methods drifted away for various problems. It has been reported that the Fragment based tacker~\cite{1640835} can follow the target by initialising at frame \#69 since it is specifically designed for handling long-term occlusion. However, from Figure~\ref{fig:exp3} one can see that the method failed to follow the target from the beginning of the sequence due to the serious scale change during frame \#1 to frame \#69. 

\subsection{Comparison on ``Car 4'' Sequence}
\begin{figure}
  \centering
  \includegraphics[width=8.5cm]{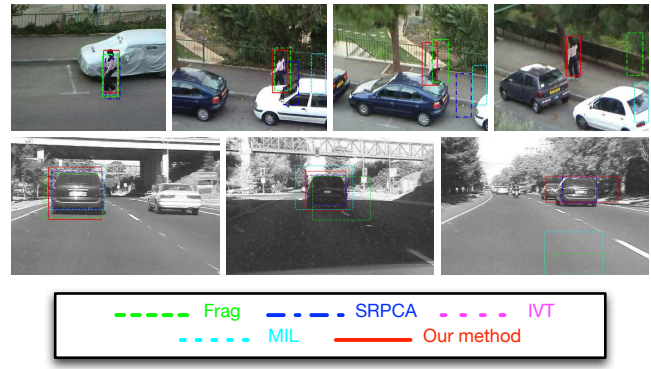}
  \caption{Qualitative tracking results over representative frames of two sequences. Images in the first row are frames \#20, \#155, \#200 and \#450 from ``Woman'' Sequence. Images in the second row are frames \#100, \#208 and \#480 from ``Car 4'' sequence.}
  \label{fig:exp3}
\end{figure}

In the last experiment, we evaluate our method on the ``Car 4'' sequence~\cite{ross2008incremental}, which contains some serious illumination and scale variation. The algorithm can process one frame of this sequence around 0.4 second. We illustrate the center errors of different methods in Figure~\ref{fig:exp2}(b). The Frag and MIL methods failed to follow the car since they are lack of effective mechanism for handling scale change. Our proposed method has no problem to track the target, however SRPCA and IVT methods show more accurate results than ours. As shown in the last instance of Figure~\ref{fig:exp3}, our method meets some trivial problems for accurately evaluating the correct component $s$ of target, which leads to small drifts of tracking results. We would like to introduce prior knowledge of component from previous frames to solve this weakness in future work.

\section{Conclusion}
In this letter, we propose a novel model based tracking method which exploits the high performance DPM in a DCRF framework. By utilising suitable temporal potential functions, the method can simultaneously handle challenging problems in tracking tasks such as variation of illumination, scale, perspective, drastic shape deformation and partial occlusion. In future work, we plan to improve the efficiency of the method with a C++ implementation. We also would like to extend current system to multiple target tracking by integrating other visual cues to discriminate targets from each other.
\ifCLASSOPTIONcaptionsoff
\newpage
\fi



\bibliographystyle{IEEEtran}
\bibliography{draft}
\end{document}